\documentclass[conference]{IEEEtran}
\IEEEoverridecommandlockouts

\usepackage{cite}
\usepackage{amsmath,amssymb,amsfonts}
\usepackage{algorithmic}
\usepackage{graphicx}
\usepackage{textcomp}
\usepackage{xcolor}
\usepackage{enumitem}
\usepackage{url}
\usepackage[utf8]{inputenc}
\usepackage[T1]{fontenc}
\usepackage{listings}
\usepackage{listingsutf8}
\usepackage{float}
\def\BibTeX{{\rm B\kern-.05em{\sc i\kern-.025em b}\kern-.08em
    T\kern-.1667em\lower.7ex\hbox{E}\kern-.125emX}}

\begin{document}

\title{Using Visual Language Models to Control Bionic Hands: Assessment of Object Perception and Grasp Inference
\thanks{The work presented in this paper was supported by the Independent Research Fund Denmark (DFF), project no. 2035-00169B ``CLIMB''.}
}

\author{Ozan Karaali$^{1}$, Hossam Farag$^{1}$, Strahinja Do\v sen$^{2}$, \v Cedomir Stefanovi\' c$^{1}$ \\
$^{1}$Department of Electronic Systems, Aalborg University, Denmark\\
$^{2}$Department of Health
Science and Technology, Aalborg University, Denmark\\
Email: \{ozank,hmf,cs\}@es.aau.dk, sdosen@hst.aau.dk
}

\maketitle

\begin{abstract}
This study examines the potential of utilizing Vision Language Models (VLMs) to improve the perceptual capabilities of semi-autonomous prosthetic hands. We introduce a unified benchmark for end-to-end perception and grasp inference, evaluating a single VLM to perform tasks that traditionally require complex pipelines with separate modules for object detection, pose estimation, and grasp planning. To establish the feasibility and current limitations of this approach, we benchmark eight contemporary VLMs on their ability to perform a unified task essential for bionic grasping. From a single static image, they should (1) identify common objects and their key properties (name, shape, orientation, and dimensions), and (2) infer appropriate grasp parameters (grasp type, wrist rotation, hand aperture, and number of fingers).
A corresponding prompt requesting a structured JSON output was employed with a dataset of 34 snapshots of common objects. Key performance metrics, including accuracy for categorical attributes (e.g., object name, shape) and errors in numerical estimates (e.g., dimensions, hand aperture), along with latency and cost, were analyzed. The results demonstrated that most models exhibited high performance in object identification and shape recognition, while accuracy in estimating dimensions and inferring optimal grasp parameters, particularly hand rotation and aperture, varied more significantly.
This work highlights the current capabilities and limitations of VLMs as advanced perceptual modules for semi-autonomous control of bionic limbs, demonstrating their potential for effective prosthetic applications.
\end{abstract}


\section{Introduction}
The human hand is an extraordinary tool, allowing for a wide variety of interactions with the environment. Replicating this functionality in prosthetic limbs remains a fundamental challenge due to limitations in current Human-Machine Interfaces (HMIs), which commonly rely on electromyography (EMG) signals that allow users to control several degrees of freedom (DOFs) sequentially ~\cite{farina_extraction_2014, chiariotti_future_2024}. This approach is slow and imposes a high cognitive load, as the burden of control is fully on the side of the user.  

Semi-autonomous prosthetic systems address user limitations through shared control between user and autonomous agent~\cite{chiariotti_future_2024, mouchoux_artificial_2021}, employing vision-based perception to analyze objects and select grasp strategies~\cite{castro_continuous_2022, chiariotti_future_2024, shi_computer_2020}. Traditional pipelines rely on YOLO for detection combined with separate modules for segmentation and pose estimation~\cite{zhang_visionbased_2025, castro_continuous_2022}. Each module needs its own development and maintenance, making the overall system complex.

Vision Language Models (VLMs) take a different approach by processing visual and textual information together~\cite{openai_gpt4_2024} in a single system. Rather than chaining multiple specialized modules, one VLM can identify objects and figure out how to grasp them, essentially serving as an "artificial exteroception" module~\cite{chiariotti_future_2024}. While general robotics frameworks like ELLMER~\cite{mon-williams_embodied_2025} and GPTArm~\cite{zhang_gptarm_2025} demonstrate this for high-level planning, the potential of VLMs for the fine-grained grasp inference needed in prosthetics remains unexplored. This simpler architecture could make prostheses easier to adapt and control, particularly for cloud-connected bionic limbs~\cite{chiariotti_future_2024}.

Despite this promise, a systematic benchmark of VLM capabilities for providing fine-grained, structured perceptual cues for bionic grasping is currently missing. This paper addresses this gap by benchmarking eight state-of-the-art VLMs on a curated dataset of 34 objects. We evaluate both object recognition and grasp inference using a unified prompt to provide the first systematic insights into their viability as perception modules for semi-autonomous prostheses.

The rest of the paper is organized as follows. Section~\ref{sec:methodology} describes our dataset, VLM testbed, prompt design, and evaluation metrics. Section~\ref{sec:results} presents the evaluation results. Section~\ref{sec:discussion} discusses the key takeaways.

\section{Methodology}
\label{sec:methodology}

\begin{figure}[!t]
    \centering
    \includegraphics[width=\columnwidth]{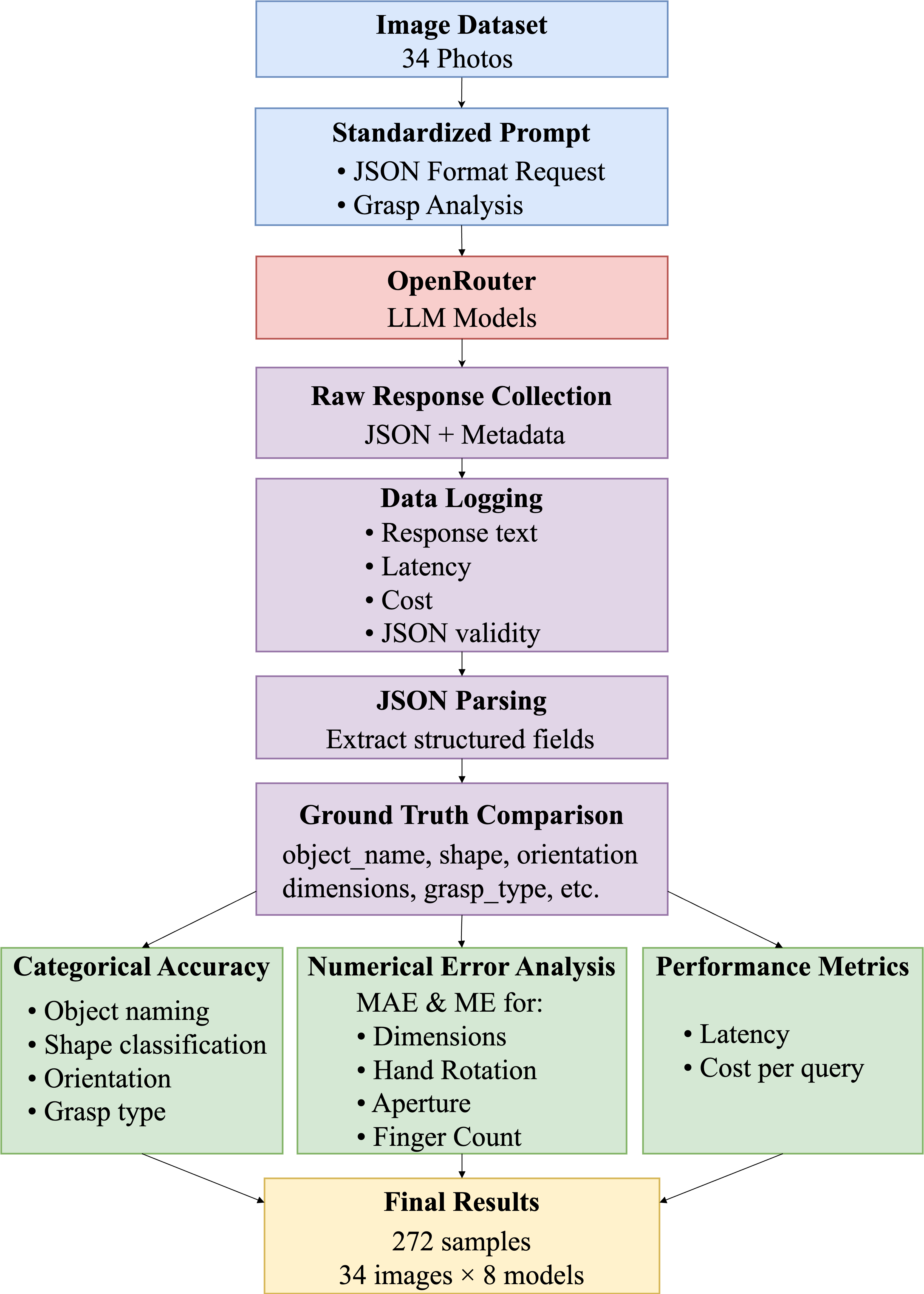}
    \caption{Diagram of the benchmarking system. Images are sent to different VLM models to estimate object properties and grasp parameters. Several categorical and numerical outcome measures are computed to assess overall performance.}
    \label{fig:benchmark}
\end{figure}

Fig.~\ref{fig:benchmark} shows the benchmarking system, which is described in the following text.

\subsection{VLM Testbed}
We evaluated eight state-of-the-art VLMs accessed via OpenRouter~\cite{openrouter_2025}: three Anthropic Claude Sonnet variants (4, 3.7, 3.5)~\cite{anthropic_claude_2025}, OpenAI GPT-4.1~\cite{openai_gpt41_2025}, Google's Gemma 3 27B and Gemini 2.5 Flash (April/May previews)~\cite{google_vlms_2025}, and Mistral Medium 3~\cite{mistral_medium3_2025}. To provide a broad and representative snapshot of the current VLM landscape, we selected these eight models from leading developers (Anthropic, OpenAI, Google, Mistral), encompassing a diverse range of architectures, training philosophies, and performance characteristics. This allows us to assess the general viability of our proposed framework rather than the capabilities of a single model.

\subsection{Image Dataset}

\begin{figure*}[!t]
    \centering
    \includegraphics[width=0.9\linewidth]{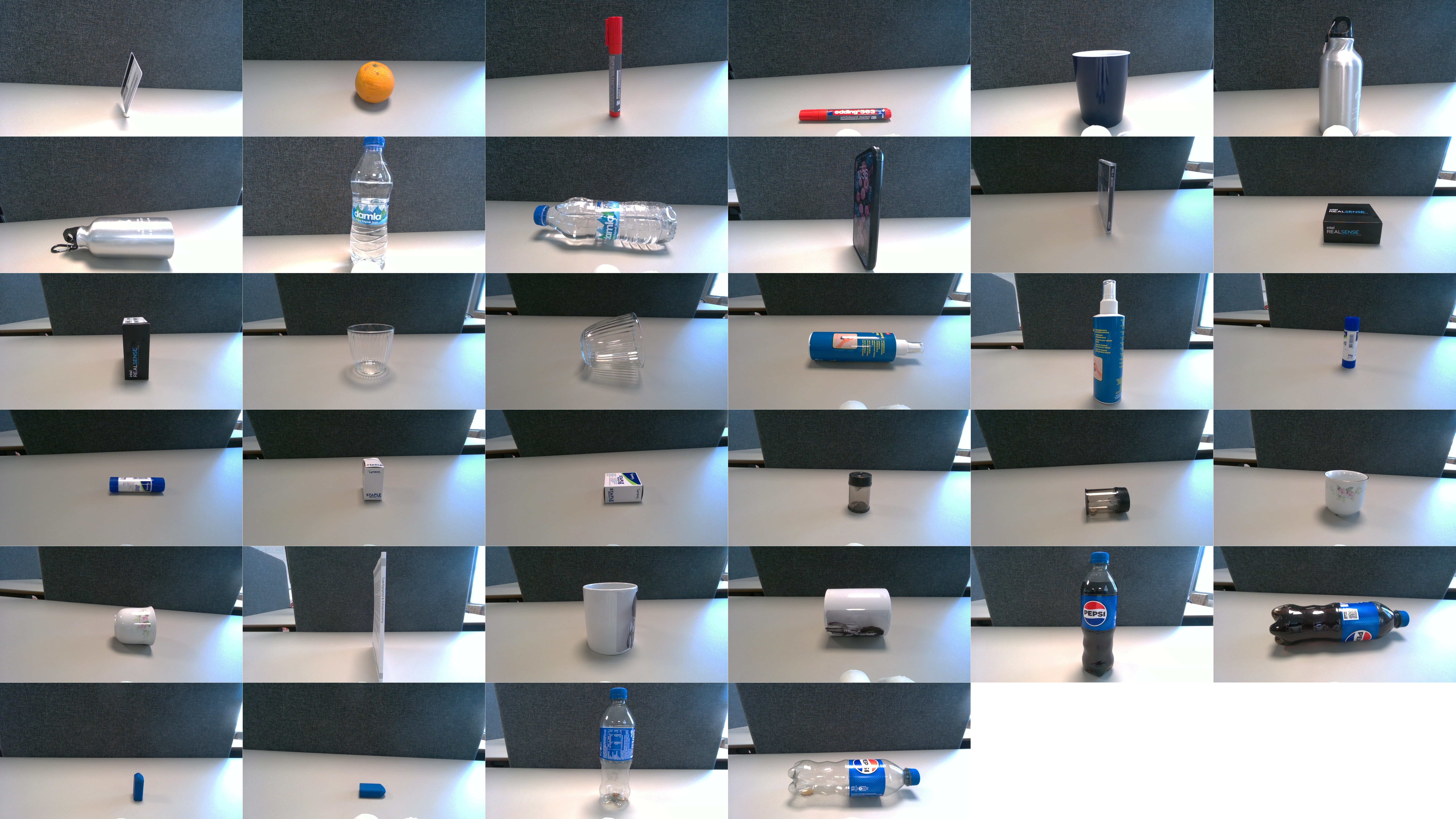}
     \caption{Image dataset of 34 common objects photographed from the perspective of an approaching prosthesis. This assumes that the camera is placed on the prosthesis, which is a common approach in the state of the art.}
    \label{fig:dataset}
\end{figure*}

The dataset comprises 34 static photographs of common household and office objects from the perspective of an approaching prosthesis (Fig.~\ref{fig:dataset}), which are selected to represent items a prosthesis user might frequently encounter. 

 The dataset includes objects of different shapes, like cylinders (23 objects, e.g., mugs, bottles), cuboids (10, e.g., boxes, books), and spheres (1, an orange); placed in vertical (19) and horizontal (15) orientations; and associated to two different grasping strategies - palmar (26, 76.5\%) and lateral (8, 23.5\%) grasp types; and varying finger counts (1-4). 
Images feature consistent lighting to focus on core visual understanding. Each image was paired with a ground truth JSON\footnote{Dataset, code, and prompts available at: \url{https://github.com/ozankaraali/benchmarking_vlms}.}.

\subsection{Prompt Design and Expected Output}
\label{ssec:prompt_design}

A single, comprehensive prompt was designed to query each VLM for each image. The question asked the VLM to identify an object in the given image and recommend how a bionic hand would grab it by acting as a robotics and computer vision expert. The VLM was explicitly asked to return its response in a structured JSON format. The key fields requested, mirroring the structure of our ground\_truth.json for each image, were:
\\
\\

\{"image\_name": "Image filename",

"object\_name": "object name",

"object\_shape": "cylinder | cuboid | sphere",

"object\_dimensions\_mm": "WxHxD",

"object\_orientation": "vertical | horizontal",

"grasp\_type": 0 (palmar) | 1 (lateral),

"hand\_rotation\_deg": "degrees (0=palm down, 90=palm sideways)",

"hand\_aperture\_mm": "est. aperture (mm)",

"num\_fingers": "number of fingers (1-4)"\}
\\
\\
The structured JSON output format enables the decomposition of complex grasp decisions into clear, benchmarkable parameters. This transparency enables the systematic evaluation of each decision component, making VLM performance measurable and trustworthy for prosthetic applications where interpretable outputs are crucial. The prompt text is listed in Appendix~\ref{app:prompt_grasp}.

\subsection{Data Collection \& Parsing}

For each image, the prompt was sent to each VLM, and the following data were logged per interaction:
\begin{itemize}
    \item Raw textual response from the VLM.
    \item Latency (total time from request to response).
    \item Cost (calculated based on token usage and model-specific pricing via OpenRouter).
    \item Validity of the JSON structure in the response.
    \item Parsed values of the fields in the expected JSON output.
\end{itemize}

\subsection{Evaluation Metrics}
The performance was evaluated using relevant metrics for both categorical and numerical data, comparing the parsed VLM outputs against the ground truth for each image.

\subsubsection*{Categorical Accuracy} 
Categorical accuracy, defined as the proportion of correct classifications out of the total number of classifications expressed in percent, was used for object names, object shape, object orientation, and grasp type. For object names, a prediction is considered correct if it matches at least one acceptable name variant (e.g., "marker," "board marker," or "board\_marker" would all be correct for a marker object). Note that for some objects, multiple grasp strategies are valid. For instance, a small box can be grasped laterally at 0 degrees or using a palmar grasp. When ground truth allowed multiple valid options (represented as a list), we computed the error using the closest valid value to the prediction.

\subsubsection*{Numerical Error} We computed Mean Absolute Error (MAE) and Mean Error (ME), where MAE indicates the average magnitude of errors and ME indicates systematic over/underestimation. We assessed object dimensions, hand rotation, hand aperture, and number of fingers. 

\section{Results}
\label{sec:results}

\subsection{Object-Level Perception}

Categorical accuracies (Table~\ref{tab:cat}) show that all models achieved high performance in object naming (82.4-97.1\%), with Mistral Medium achieving the highest accuracy. For shape recognition, six models achieved perfect 100\% accuracy, while the remaining two exceeded 97\%. Object orientation was also recognized with high accuracy (85.3-100\%), with Claude 3.5 and 3.7 Sonnet achieving perfect performance.

Grasp type classification showed more variation across models, ranging from 91.2\% (Gemini Flash 17-04) down to 44.1\% (Gemma 3).
Some models predicted palmar grasps for slender objects, such as markers, which is technically valid but less efficient than a lateral grasp.

\begin{table}[t]
\caption{Categorical accuracies (\%); the best (\textcolor{blue}{\textbf{bold}}) and worst (\textcolor{red}{\textit{italic}}) results are highlighted.}
\begin{center}
\begin{tabular}{|l|c|c|c|c|}
\hline
\textbf{Model} & \textbf{Naming} & \textbf{Shape} & \textbf{Orient.} & \textbf{Grasp type} \\
\hline
Claude 3.5 Sonnet                   & 91.2 & \color{blue}\textbf{100.0} & \color{blue}\textbf{100.0} & 85.3 \\
Claude 3.7 Sonnet                   & 91.2 & \color{blue}\textbf{100.0} & \color{blue}\textbf{100.0} & 85.3 \\
Claude Sonnet 4                     & \color{red}\textit{82.4} & \color{red}\textit{97.1} & 97.1 & 79.4 \\
GPT-4.1                             & 94.1 & \color{blue}\textbf{100.0} & 97.1 & 82.4 \\
Gemini 2.5 Fl. 17-04                & 91.2 & \color{blue}\textbf{100.0} & 97.1 & \color{blue}\textbf{91.2} \\
Gemini 2.5 Fl. 20-05                & 88.2 & \color{blue}\textbf{100.0} & 97.1 & 82.4 \\
Gemma 3 27B-IT                      & 88.2 & \color{blue}\textbf{100.0} & 91.2 & \color{red}\textit{44.1} \\
Mistral Medium                      & \color{blue}\textbf{97.1} & \color{red}\textit{97.1} & \color{red}\textit{85.3} & 58.8 \\
\hline
\end{tabular}
\label{tab:cat}
\end{center}
\end{table}

\subsection{Object size and distance perception}

 Table~\ref{tab:dim} shows the models' ability to estimate object dimensions from single images. Gemini 2.5 Flash 20-05 achieved the lowest MAE across all dimensions, with particularly strong performance in depth estimation (4.62 ± 4.06 mm).

\begin{table}[t]
\caption{MAE ± STD for object dimensions (mm); the lowest (\textcolor{blue}{\textbf{bold}}) and highest (\textcolor{red}{\textit{italic}}) MAE are highlighted.}
\begin{center}
\begin{tabular}{|l|c|c|c|}
\hline
\textbf{Model} & \textbf{Width} & \textbf{Height} & \textbf{Depth} \\
\hline
Claude 3.5 Sonnet                   & \color{red}\textit{10.15 ± 10.44} & 16.03 ± 18.02 & 6.35 ± 5.97 \\
Claude 3.7 Sonnet                   & 8.91 ± 10.24 & 14.94 ± 18.71 & 5.97 ± 5.27 \\
Claude Sonnet 4                     & 8.56 ± 12.55 & 15.88 ± 16.73 & 5.71 ± 8.14 \\
GPT-4.1                             & 8.32 ± 10.60 & 12.12 ± 13.65 & 5.74 ± 6.35 \\
Gemini 2.5 Fl. 17-04                & 9.35 ± 10.91 & 13.71 ± 16.60 & 6.24 ± 5.39 \\
Gemini 2.5 Fl. 20-05                & \color{blue}\textbf{7.03 ± 10.35} & \color{blue}\textbf{8.59 ± 9.79} & \color{blue}\textbf{4.62 ± 4.06} \\
Gemma 3 27B-IT                      & 9.53 ± 7.67 & \color{red}\textit{19.44 ± 21.46} & \color{red}\textit{8.71 ± 7.33} \\
Mistral Medium                      & 7.71 ± 9.49 & 10.50 ± 10.22 & 6.47 ± 6.92 \\
\hline
\end{tabular}
\label{tab:dim}
\end{center}
\end{table}

\subsection{Estimation of hand rotation and aperture}

Table~\ref{tab:grasp} presents the accuracy of grasp parameter estimation. Hand rotation showed the highest variability among all parameters, with MAEs ranging from 13.2° to 47.6°. Claude 3.5 Sonnet and Mistral Medium achieved the lowest MAE for finger count estimation (0.50), while GPT-4.1 showed the best hand aperture estimation (5.74 ± 6.14 mm).

\begin{table}[t]
\caption{MAE ± STD for grasp parameters; the lowest (\textcolor{blue}{\textbf{bold}}) and highest (\textcolor{red}{\textit{italic}}) MAE are highlighted.}
\begin{center}
\begin{tabular}{|l|c|c|c|}
\hline
\textbf{Model} & \textbf{Rot. (°)} & \textbf{Apert. (mm)} & \textbf{Fingers} \\
\hline
Claude 3.5 Sonnet                   & 21.2 ± 38.2 & 9.79 ± 12.77 & \color{blue}\textbf{0.50 ± 0.88} \\
Claude 3.7 Sonnet                   & 34.4 ± 43.7 & 6.15 ± 5.64 & 0.62 ± 0.73 \\
Claude Sonnet 4                     & 14.6 ± 32.2 & 9.79 ± 10.84 & \color{red}\textit{0.88 ± 0.87} \\
GPT-4.1                             & \color{red}\textit{47.6 ± 44.9} & \color{blue}\textbf{5.74 ± 6.14} & 0.53 ± 0.78 \\
Gemini 2.5 Fl. 17-04                & 45.6 ± 62.4 & 14.18 ± 23.37 & 0.56 ± 0.95 \\
Gemini 2.5 Fl. 20-05                & 26.5 ± 41.0 & 7.32 ± 7.40 & 0.62 ± 1.06 \\
Gemma 3 27B-IT                      & \color{blue}\textbf{13.2 ± 31.9} & \color{red}\textit{16.12 ± 19.39} & 0.74 ± 1.01 \\
Mistral Medium                      & 37.1 ± 44.3 & 9.94 ± 11.02 & \color{blue}\textbf{0.50 ± 0.74} \\
\hline
\end{tabular}
\label{tab:grasp}
\end{center}
\end{table}

\subsection{Bias Analysis (Mean Error)}
Tables~\ref{tab:dim_me} and \ref{tab:grasp_me} show the ME analysis, revealing systematic biases in model predictions. A ME close to 0 indicates unbiased predictions, while positive and negative values indicate overestimation and underestimation, respectively.

\begin{table}[t]
\caption{ME ± STD for object dimensions (mm); ME closest to 0 (\textcolor{blue}{\textbf{bold}}) and furthest (\textcolor{red}{\textit{italic}}) are highlighted.}
\begin{center}
\begin{tabular}{|l|c|c|c|}
\hline
\textbf{Model} & \textbf{Width} & \textbf{Height} & \textbf{Depth} \\
\hline
Claude 3.5 Sonnet                   & 0.62 ± 14.55 & 12.09 ± 20.87 & 2.35 ± 8.39 \\
Claude 3.7 Sonnet                   & 1.38 ± 13.50 & 9.88 ± 21.81 & 2.26 ± 7.63 \\
Claude Sonnet 4                     & \color{blue}\textbf{-0.26 ± 15.19} & 5.12 ± 22.49 & 2.41 ± 9.64 \\
GPT-4.1                             & 1.21 ± 13.43 & 7.65 ± 16.57 & 4.15 ± 7.49 \\
Gemini 2.5 Fl. 17-04                & \color{red}\textit{-7.00 ± 12.55} & 4.82 ± 20.98 & -2.88 ± 7.72 \\
Gemini 2.5 Fl. 20-05                & -4.38 ± 11.71 & 2.35 ± 12.81 & \color{blue}\textbf{-0.97 ± 6.07} \\
Gemma 3 27B-IT                      & 2.12 ± 12.05 & \color{red}\textit{15.26 ± 24.61} & \color{red}\textit{4.18 ± 10.59} \\
Mistral Medium                      & -2.18 ± 12.03 & \color{blue}\textbf{0.44 ± 14.65} & -1.47 ± 9.36 \\
\hline
\end{tabular}
\label{tab:dim_me}
\end{center}
\end{table}

For grasp parameters, GPT-4.1 and Claude 3.7 Sonnet showed minimal bias in aperture estimation (0.44 ± 8.39 mm and 0.15 ± 8.34 mm, respectively), which is favorable for practical use.

\begin{table}[t]
\caption{ME ± STD for grasp parameters; ME closest to 0 (\textcolor{blue}{\textbf{bold}}) and furthest (\textcolor{red}{\textit{italic}}) are highlighted.}
\begin{center}
\begin{tabular}{|l|c|c|c|}
\hline
\textbf{Model} & \textbf{Rot. (°)} & \textbf{Apert. (mm)} & \textbf{Fingers} \\
\hline
Claude 3.5 Sonnet                   & 21.18 ± 38.18 & -0.97 ± 16.06 & 0.21 ± 0.99 \\
Claude 3.7 Sonnet                   & \color{red}\textit{34.41 ± 43.74} & \color{blue}\textbf{0.15 ± 8.34} & \color{blue}\textbf{-0.09 ± 0.95} \\
Claude Sonnet 4                     & \color{blue}\textbf{-1.32 ± 35.34} & 5.44 ± 13.56 & \color{red}\textit{-0.71 ± 1.02} \\
GPT-4.1                             & 5.29 ± 65.27 & 0.44 ± 8.39 & -0.41 ± 0.84 \\
Gemini 2.5 Fl. 17-04                & 28.53 ± 71.79 & -3.24 ± 27.14 & 0.26 ± 1.07 \\
Gemini 2.5 Fl. 20-05                & -10.59 ± 47.65 & -4.26 ± 9.50 & 0.38 ± 1.16 \\
Gemma 3 27B-IT                      & 7.94 ± 33.59 & \color{red}\textit{10.59 ± 22.89} & 0.32 ± 1.21 \\
Mistral Medium                      & 15.88 ± 55.53 & -3.24 ± 14.49 & -0.32 ± 0.83 \\
\hline
\end{tabular}
\label{tab:grasp_me}
\end{center}
\end{table}

\subsection{Latency and Cost}

Table~\ref{tab:latency} presents practical deployment considerations. Gemini 2.5 Flash 20-05 achieved the fastest response time (2.60 ± 0.47 s), while Gemma 3 27B-IT offered the lowest cost per query (\$0.00016). The Claude models showed the highest latencies (7-9 s) and costs (\$0.012/query).

\begin{table}[t]
\caption{Mean latency ± STD (s) and cost per query; best (\textcolor{blue}{\textbf{bold}}) and worst (\textcolor{red}{\textit{italic}}) values are highlighted.}
\begin{center}
\begin{tabular}{|l|c|c|}
\hline
\textbf{Model} & \textbf{Latency (s)} & \textbf{Cost (\$)} \\
\hline
Claude 3.5 Sonnet                   & 7.14 ± 1.34 & 0.01215 \\
Claude 3.7 Sonnet                   & \color{red}\textit{9.41 ± 3.61} & 0.01214 \\
Claude Sonnet 4                     & 7.70 ± 1.33 & \color{red}\textit{0.01215} \\
GPT-4.1                             & 6.71 ± 1.64 & 0.00425 \\
Gemini 2.5 Fl. 17-04                & 3.26 ± 0.47 & 0.00106 \\
Gemini 2.5 Fl. 20-05                & \color{blue}\textbf{2.60 ± 0.47} & 0.00106 \\
Gemma 3 27B-IT                      & 6.53 ± 3.96 & \color{blue}\textbf{0.00016} \\
Mistral Medium                      & 3.63 ± 1.29 & 0.00097 \\
\hline
\end{tabular}
\label{tab:latency}
\end{center}
\end{table}

\section{Discussion}
\label{sec:discussion}

The presented results highlight both strengths and shortcomings, and a set of trade-offs that are discussed below.

\subsection{Categorical accuracy}

\textbf{Categorical object properties can be estimated with high accuracy.}  
As Table \ref{tab:cat} shows, every model exceeded 88\% accuracy for object naming, shape, and orientation, with four models achieving perfect performance on shape and two on orientation. This means that today's VLMs can already deliver reliable symbolic labels for individual, well-lit objects, which is an essential prerequisite for downstream grasp planning. Specifically, the shape can be used to determine grasp type, whereas the pose determines the hand orientation (wrist angle).

\textbf{Grasp type estimation accuracy varies significantly across models.} While the best-performing model (Gemini Flash 17-04) achieved 91.2\% accuracy, others performed substantially worse, with Gemma 3 achieving only 44.1\%. Because grasp type is a binary choice (palmar vs lateral), errors mean the hand would perform the wrong grasp type. While this can result in suboptimal strategies (e.g., grasping a bottle using a lateral grasp when a palmar would be more stable), in many cases, it would not prevent successful grasping. Still, the results suggest room for improvement through task-specific fine-tuning or interactive confirmation steps.

\textbf{Structured JSON improves interpretability.}  
All 272 model-image pairs produced syntactically valid JSON, i.e., the needed parameters (shape, size, aperture, etc.) arrived correctly parsed. Therefore, the prosthesis controller can read the output of the VLMs analysis and use the parsed information to preshape the hand accordingly.

\subsection{Numerical Accuracy and Consistency}
The breakdown of the quantitative results reveals nuances in both average error and prediction consistency.

\textbf{Dimensions.} All Mean Absolute Errors are below 20 mm; for width and depth, they are typically below 11 mm (Table \ref{tab:dim}). On an 80 mm-wide mug, an 11 mm MAE equals a 13.8\% error-visible, yet within the 15\% tolerance we define as acceptable for initial hand shaping. Standard deviations qualify this further: Gemini 2.5 Flash 20-05 achieves a depth MAE of 4.62 mm ± 4.06 mm, whereas its width MAE of 7.03 mm comes with a larger spread ± 10.35 mm, suggesting that the analysis of some images would lead to prosthesis hand preshapes that may still need user correction.

The Mean Error analysis reveals systematic biases crucial for practical implementation. A positive ME means the hand is preset wider than necessary, which is preferable as the user can simply continue closing. A negative ME implies insufficient hand opening, forcing extra user effort to reopen before grasping. For example, GPT-4.1 and Claude 3.7 Sonnet exhibit near-zero aperture ME (0.44 mm and 0.15 mm, respectively) with relatively low variability (±8.39 mm and ±8.34 mm), representing ideal characteristics for practical use.

\textbf{Hand rotation.} Gemma 3 posts the lowest MAE for rotation (13.2°, Table \ref{tab:grasp}). However, the accompanying high STD (±31.9°) suggests this low average error might result from frequently correct predictions for canonical orientations mixed with occasional large errors. Importantly, hand rotation can be determined indirectly from object orientation and grasp type, both of which can be estimated with high accuracy. In general, relatively large errors in rotation estimation can be tolerated since the user can correct this through compensatory motions (e.g., rotating the prosthesis from the shoulder to compensate for an incorrect orientation of the hand). 

\textbf{Aperture and fingers.} GPT-4.1 estimates aperture with the lowest MAE (5.74 mm), though the ±6.14 mm STD indicates considerable variability. For finger count, classification accuracy varied significantly (Table \ref{tab:grasp}), highlighting inconsistencies in this discrete estimation task.

\subsection{Latency-Cost Trade-offs}
Gemini Flash 20-05 is the fastest (2.6 s), while Gemma 3 is the cheapest (\$0.00016). The Claude Sonnets featured latencies of 7-9~s and costs of ~\$0.012/query, indicating a significant challenge for real-time prosthetic control. Traditional systems operate at 8 Hz (150 ms processing)~\cite{castro_continuous_2022}, while complete grasping sequences take 6-14~s. A 2-3~s delay from the fastest models might be tolerable if processing occurs during reaching, but 7-9~s delays would frustrate users accustomed to immediate response.

\subsection{Implications for Prosthetics}

The benchmark results indicate that VLMs can reliably infer high-level categorical properties (e.g., shape, orientation), but still exhibit variability in estimating precise grasp parameters. These findings support the use of shared control, where a VLM provides an interpretable, JSON-structured initial guess, and the user or a lightweight controller makes adjustments as needed. This approach leverages VLMs' strengths in object understanding while compensating for their limitations in precise parameter estimation. The structured JSON output facilitates the shared control paradigm by making mismatches (e.g., predicted vs. actual aperture) easy to spot and correct. 

Future work could explore adaptive, closed-loop learning via user feedback to improve model performance over time. In addition, future work could explore incorporating domain knowledge directly into prompts, such as the relationship between object orientation and hand rotation, which could potentially improve performance in challenging subtasks.

\subsection{Limitations}

While this study provides a comprehensive benchmark of VLMs for prosthetic perception and grasp inference, several key limitations must be acknowledged.

The evaluation was conducted on a relatively small dataset of 34 objects in single-object scenes under controlled lighting conditions. The dataset distribution, with 23 cylindrical objects, 10 cuboids, and only 1 sphere, introduces an inherent imbalance that may limit the generality of the conclusions. Specifically, the prevalence of cylindrical objects may mean the reported accuracy for dimension and grasp estimation is more representative for that shape class than for others. 
Real-world scenarios involving clutter, occlusion, varying illumination, and multiple objects remain to be tested. Future work should validate these findings on a larger dataset with balanced shape distribution and more complex scenes.

The grasp strategies used to generate ground truth are inherently subjective. Many objects admit multiple valid grasp approaches, and what constitutes an "optimal" grasp depends on task requirements, user preferences, and biomechanical constraints beyond our current evaluation framework.

All tested VLMs were evaluated in a zero-shot setting without task-specific fine-tuning. Domain-specific training on prosthetic grasping datasets would likely improve performance significantly, particularly for grasp parameter estimation, where current errors remain substantial.

The VLM inference latencies (ranging from 2.6 to 9.4~s) are too long for seamless real-time control compared to traditional vision systems. While users might tolerate delays if they provide superior functionality, further optimization will be necessary for practical deployment. This could involve edge deployment, model compression, or the use of specialized, smaller VLMs designed for faster inference.

Finally, our benchmark focuses on static object analysis rather than dynamic grasping scenarios. Real prosthetic control involves temporal considerations, user feedback integration, and adaptive control that this evaluation does not capture. Nevertheless, the present paper shows that this is a promising area of research, potentially leading to a new generation of bionic limbs that can leverage powerful computer vision resources available through cloud/edge computing. 

\bibliographystyle{IEEEtran}
\bibliography{references} 

\appendices           

\section{Prompt - Grasp Parameter Inference}

\label{app:prompt_grasp}


\begin{lstlisting}[basicstyle=\small\ttfamily,breaklines=true]
You are an expert roboticist controlling a LEFT bionic hand with a camera on top. You will see ONE image.
Start pose: palm down, thumb pointing right.

Task
----
Identify the main object in the image and derive its properties and suitable grasp parameters for the LEFT bionic hand.

Return **EXACTLY one line of JSON**--no commentary, no line breaks:
{"object_name":"<object_name>","object_shape":"<cuboid|cylinder|sphere>","object_dimensions_mm":"<W>x<H>x<D>","object_orientation":"<horizontal|vertical>","grasp_type":<0|1>,"hand_rotation_deg":<float>,"hand_aperture_mm":<float>,"num_fingers":<int>}

Parameter Rules and Clarifications
----------------------------------
**Object Properties:**
• `object_name`: Generic name of the main identified object (e.g., "bottle", "book", "screwdriver").
• `object_shape`: The primary geometric shape of the main object. Choose from: "cuboid", "cylinder", "sphere".
• `object_dimensions_mm`: Estimated dimensions of the object in millimeters, formatted as a string "<W>x<H>x<D>" (e.g., "150x80x30").
    • W (width) = second-longest side of the object.
    • H (height) = longest side of the object.
    • D (depth) = shortest side of the object.
• `object_orientation`: The dominant orientation of the object as it appears in the image. Choose from: "horizontal", "vertical".

**Grasp Parameters (for the LEFT bionic hand):**
• `grasp_type`: Specify the type of grasp.
    • 0 = Palmar grasp (thumb opposes the palm side of the middle and/or ring fingers).
    • 1 = Lateral grasp (thumb opposes the radial side of the index finger, like holding a key or another thin object).
• `hand_rotation_deg`: Rotation of the hand in degrees, relative to the start pose (palm down, thumb pointing right), to appropriately grasp the object.
    • 0 = Palm remains down.
    • +90 = Palm rotates towards the left (handshake position, thumb points upwards).
    • +180 = Palm rotates to face upwards (thumb points towards the left).
    • -90 = Palm rotates towards the right (thumb points downwards).
    Interpolate for intermediate angles if necessary. Keep in mind that lateral grasp is done perpendicular to the object, to secure the object in between thumb and index finger.
• `hand_aperture_mm`: Estimate the required opening between the thumb and fingers in millimeters to encompass and securely hold the object. This should be a reasonable value for the object's size.
• `num_fingers`: The number of fingers (1, 2, 3, or 4) to use for the palmar grasp, where lateral grasp always uses 1, thumb only, excluding the thumb (the thumb is always used). Estimate this based on the height of the object's surface that the fingers will make contact with:
    • 1 finger: for object contact height up to 10mm covered by fingers.
    • 2 fingers: for object contact height >10mm and up to 25mm covered by fingers.
    • 3 fingers: for object contact height >25mm and up to 40mm covered by fingers.
    • 4 fingers: for object contact height >50mm covered by fingers.
    Always prioritize a stable grasp configuration appropriate for the object's shape and estimated size.

Example
-------
{"object_name":"mug","object_shape":"cylinder","object_dimensions_mm":"80x95x80","object_orientation":"vertical","grasp_type":1,"hand_rotation_deg":90,"hand_aperture_mm":75,"num_fingers":4}
\end{lstlisting}

\end{document}